\documentclass[letterpaper, 10 pt, journal, twoside]{IEEEtran}

\IEEEoverridecommandlockouts                              

\usepackage{amsmath} 
\usepackage{amssymb}  
\usepackage{color}
\usepackage{graphicx}
\usepackage{graphics}
\usepackage{caption}    
\usepackage{subcaption}
\usepackage{multirow}
\usepackage{soul}
\usepackage{booktabs}
\usepackage{verbatim}
\usepackage{tabularx}

\usepackage[noend]{algpseudocode}

\usepackage{pifont}

\usepackage[mathscr]{eucal}
\usepackage{amssymb}
\usepackage{amsmath}
\usepackage[nolist]{acronym} 

\definecolor{red}{RGB}{255, 0, 0}
\definecolor{black}{RGB}{0, 0, 0}
\definecolor{orange}{RGB}{252, 130, 62}
\definecolor{blue}{RGB}{0, 0,255}

\newcommand{\new}{\color{black}} 
\newcommand{\black}{\color{black}} 

\usepackage{booktabs}
\usepackage[usenames,dvipsnames]{xcolor} 
\definecolor{lavender}{rgb}{0.9, 0.9, 0.98}

\usepackage[font=small,labelfont=bf]{caption}
\usepackage{colortbl}
\newacro{ea}[EA]{Exploratory Action}
\newacro{gnn}[GNN]{Graph Neural Network}
\newacro{pcd}[PCD]{point cloud}
\newacro{mlp}[MLP]{Multi-Layer Perceptron}
\newacro{rnn}[RNN]{Recurrent Neural Network}
\newacro{pp}[PP]{Physical Properties}

\usepackage{afterpage}
\usepackage[linesnumbered,ruled,vlined]{algorithm2e}
\SetKwInput{KwInput}{Input}                
\SetKwInput{KwOutput}{Output}              
\usepackage{hyperref}
\usepackage{multicol}
\usepackage{soul}
\usepackage{url}

\begin{document}

\title{
AdaFold: Adapting Folding Trajectories of Cloths\\ via Feedback-loop Manipulation
}

\author{Alberta Longhini${^{1}}$, Michael C. Welle${^1}$, Zackory Erickson${^2}$, and Danica Kragic${^1}$
\thanks{Manuscript received: March, 9th, 2024; Revised June, 12th, 2024; Accepted July, 7th, 2024.}
\thanks{This paper was recommended for publication by Editor Abhinav Valada upon evaluation of the Associate Editor and Reviewers' comments.
This work was supported by the Swedish Research Council, Knut and Alice Wallenberg Foundation, and the European Research  Council (ERC-884807).}
\thanks{$^{1}$The authors are with the Robotics, Perception and Learning Lab, EECS, at KTH Royal Institute of Technology, Stockholm, Sweden
     {\tt\small albertal, mwelle, dani@kth.se}}%
        \thanks{    $^{2}$The authors are with are with Carnegie Mellon University, Pittsburgh, USA
        {\tt\small  zerickso@andrew.cmu.edu}}%
        \thanks{Digital Object Identifier (DOI): see top of this page.}
}


\markboth{IEEE Robotics and Automation Letters. Preprint Version. Accepted July, 2024}
{Longhini \MakeLowercase{\textit{et al.}}: AdaFold: Adapting Folding Trajectories of Cloths via Feedback-loop Manipulation}

\maketitle

\begin{abstract}
We present AdaFold, a model-based feedback-loop framework for optimizing folding trajectories.  AdaFold extracts a particle-based representation of cloth from RGB-D images and feeds back the representation to a model predictive control to re-plan folding trajectory at every time-step. A key component of AdaFold that enables feedback-loop manipulation is the use of semantic descriptors extracted from \new{geometric features.}
\black{These descriptors enhance the particle representation of the cloth to distinguish between ambiguous point clouds of differently folded cloths. Our experiments demonstrate AdaFold's ability to adapt folding trajectories of cloths with varying physical properties and generalize from simulated training to real-world execution. Videos can be found on our project website: \href{https://adafold.github.io/}{https://adafold.github.io}}
\end{abstract}


\section{Introduction}

\IEEEPARstart{G}{eneralizing} robotic manipulation skills requires adapting to object variations such as pose, shape, and physical properties \cite{kroemer2021review}.
Feedback-loop manipulation represents a class of methods to adapt to these variations. The effectiveness of this class of methods, however, is heavily contingent upon the robot's ability to accurately perceive and track the state of the object throughout the manipulation. 
Within the realm of deformable objects, such as cloth, feedback-loop manipulation remains under-explored due to challenges in state estimation and dynamics modelling \cite{zhu2022challenges,longhini2024unfolding}.

Recent advances in learning cloth dynamics have led to model-based methods to plan pick-and-place interactions for folding and flattening \cite{lin2022learning,ma2021learning}. 
Despite the progress, these methods rely on open-loop planning and pre-defined manipulation trajectories due to the practical challenges of estimating and tracking cloth states during manipulation. 
A promising approach to mitigate these challenges is integrating semantic knowledge {\new based on geometric features}~\cite{miller2012geometric} 
into potentially ambiguous, but easy to track, representations such as point clouds.  

\begin{figure}[t!]
    \centering
    \includegraphics[width=0.45\textwidth]{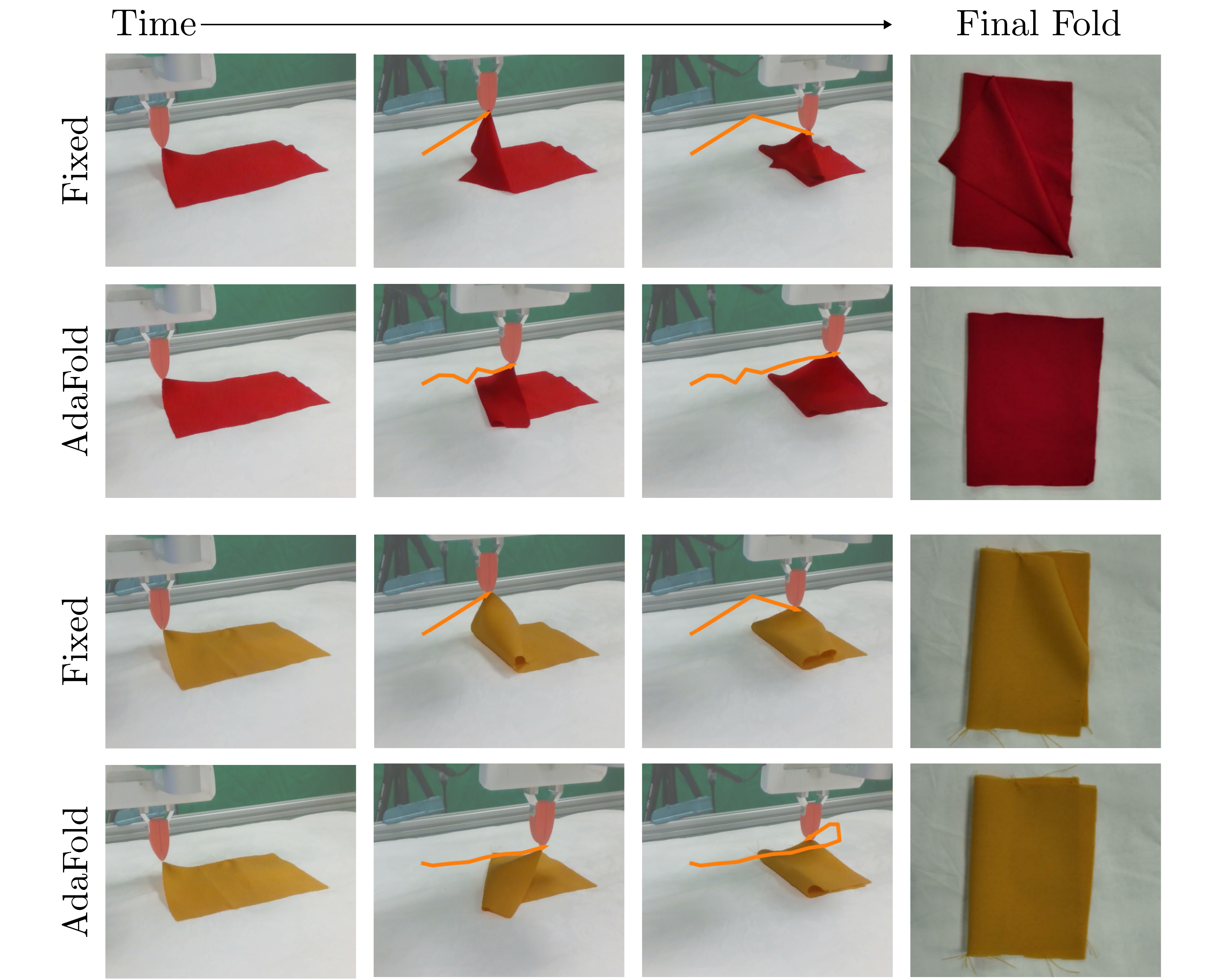}
    \caption{ \textbf{AdaFold} successfully adapts the folding trajectories of the two cloths with different physical properties, achieving a better folding than a predefined triangular trajectory. }
    \label{fig:intro_example}
     \vspace{-0.2in}
\end{figure}

\begin{figure*}[t]
\centering
\includegraphics[width=\linewidth]{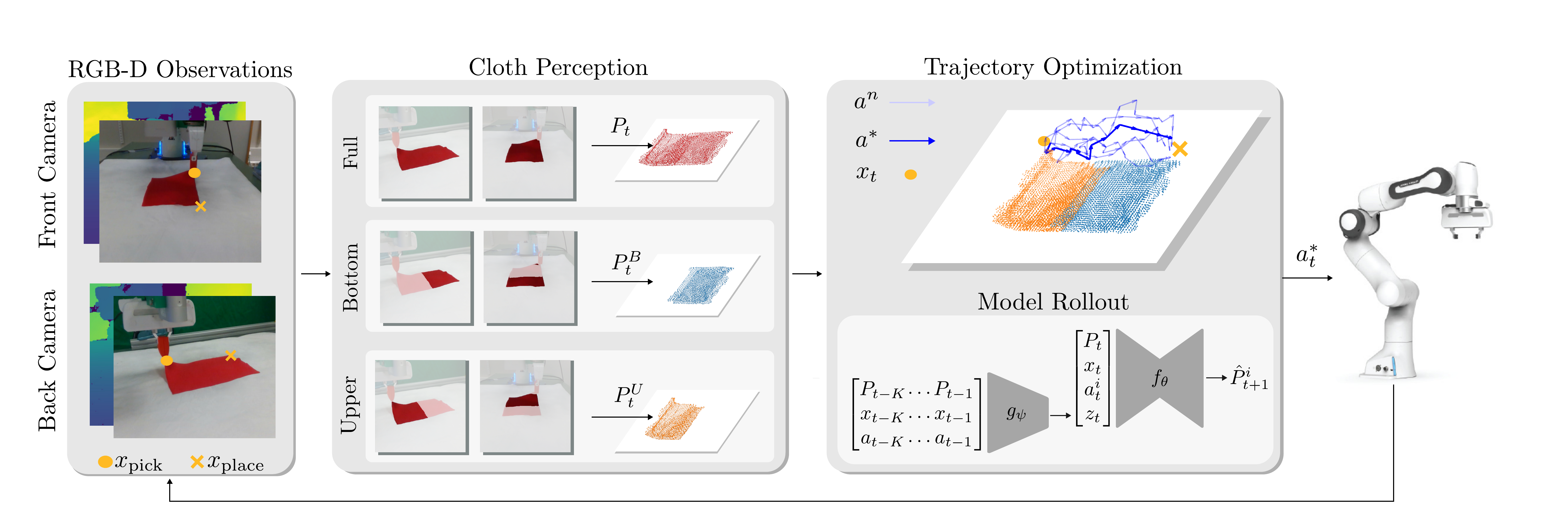}

\caption{\textbf{Overview of AdaFold for feedback-loop manipulation of cloths.} Given a set of pick-and-place positions  $(x_{pick}, x_{place})$, AdaFold optimizes the best folding action  $a^*_t$ at each time-step $t$. RGB-D observations from different calibrated cameras are used to extract point cloud representations with semantic descriptors. The semantic descriptors \textit{Upper} and \textit{Bottom} are obtained based on geometric features following~\cite{miller2011parametrized}.  The optimal folding action $a^*_t$ is obtained with MPC, which uses the forward and adaptation modules $f_\theta$ and $g_\psi$ to evaluate the candidate trajectories $a^n$ (light blue) and update the optimal control sequence $a^*$ (dark blue).
}
\label{fig:method_all}
  \vspace{-0.15in}
\end{figure*}

We propose AdaFold, a model-based framework for feedback-loop manipulation of cloth to optimize folding trajectories. AdaFold relies on particle-based state representation and a learned model of the cloth to optimize the best sequence of folding actions with model-predictive control (MPC)~\cite{camacho2013model}. To adapt to cloth variations, we perform the manipulation in a feedback-loop fashion by re-planning the folding trajectory after every time-step. We further propose to use {\new parameterized shape models for cloths~\cite{miller2011parametrized}} 
to extract semantic descriptors of the \textit{upper} and the \textit{bottom} layers of the cloth from RGB images. To overcome the challenge of {\new tracking the upper and bottom 
layers through manipulation, we leverage recent advances in video tracking~\cite{cheng2022xmem} to track the respective masks.} 

We evaluate AdaFold both in simulation and in the real world using a single-arm manipulator to perform the half-folding task proposed in \cite{garcia2020benchmarking}. The results confirm the capability of AdaFold to optimize the folding trajectory of both simulated and real-world cloths, successfully accounting for variations in physical properties, demonstrated in Fig.\ref{fig:intro_example}, initial position and size of the cloth. 
We further assess the benefit of introducing semantic descriptors into the state representation of the cloth to disambiguate similar point cloud observations of different folded states. 
In summary, our contributions are:
\begin{itemize}
    \item A model-based approach to optimize the folding trajectory in a feedback-loop fashion, which transfers to real-world cloths with unknown physical properties.
    \item A method to embed semantic descriptors into a point cloud representation of cloth by leveraging {\new geometric features along pre-trained video tracking techniques}.
    \item An extensive evaluation in both simulation and real-world environments, considering variations of object properties such as pose, size and physical properties.
\end{itemize}

\section{Related Work}

\subsection{State Representation}
{\new Geometric features, such as corners or landmark points, are effective representations when the cloth lies flat on a surface, but fail for more crumpled cloth configurations~\cite{moletta2022representing}.}
While image-based representations alleviate the need for explicit state estimation and tracking~\cite{yan2021learning,hoque2022visuospatial}, they are sensitive to variations in color, brightness, and camera perspective. 
Particle-based representations, such as graphs, are less sensitive to visual clues and achieve better generalization to novel cloth shapes and textures \cite{ma2021learning,lin2022learning}. Yet, tracking graph representations is still a challenge \cite{huang2023self}. Hierarchical and bottom-up representations \cite{popovic2011grasping} have not yet been demonstrated in the context of cloth manipulation. 

Point clouds, on the other hand, demand less computational effort for perception tasks due to their unstructured nature and have shown success in assistive dressing tasks \cite{wang2023one}. Nevertheless, in situations with significant self-occlusions, point cloud representations become ambiguous as different layers of the cloth cannot be distinguished based solely on the observable set of points. While classical computer vision approaches such as a Harris Corner Detector \cite{willimon2011model} or a wrinkle-detector \cite{qiu2023robotic} can be used for detecting cloth features, they are typically not robust to variations of texture, lightning conditions, and non-static observations. This study tackles these perception challenges by augmenting point cloud representations with semantic descriptors, which we derive from RGB observations through {\new geometric features}.

\subsection{Cloth manipulation} 
Heuristic-based methods for cloth manipulation are viable approaches for cloth folding and flattening~\cite{miller2012geometric}, although they may struggle to generalize across diverse real-world scenarios. Learning-based methods emerge as an alternative when heuristics fall short, and can be divided into two categories: model-free and model-based. 
Model-free learning is a class of methods that circumvents the challenges in state estimation and dynamics modeling {\new while also scaling to the real world}. This technique has been successfully used to tackle several cloth manipulation tasks, such as folding or flattening, by finding the best sequence of pick-and-place positions \cite{
seita2020deep,wu2019learning,mo2023learning,wang2023one,petrik2019feedback}.  More recently, \cite{hietala2022learning} proposed a model-free visual feedback policy to fold cloths in half, successfully adapting the manipulation trajectory to three real-world cloths. To address the sample inefficiency of model-free learning \cite{duan2016benchmarking}, the authors relied on human expert demonstrations. Model-based techniques, on the other hand, have shown promise in sample-efficient learning \cite{williams2017information, lin2022learning,ma2021learning}. These approaches typically employ a top-down camera view to minimize self-occlusions of the cloth, which, however, restricts visibility of the cloth's intermediate states during manipulation. Consequently, this limitation hinders the use of feedback-loop strategies. In this work, we address this limitation by leveraging side camera views to reduce occlusions and propose a model-based approach specifically designed for feedback-loop manipulation.

\begin{figure*}[t]
    \centering
    \includegraphics[width=0.9\textwidth]{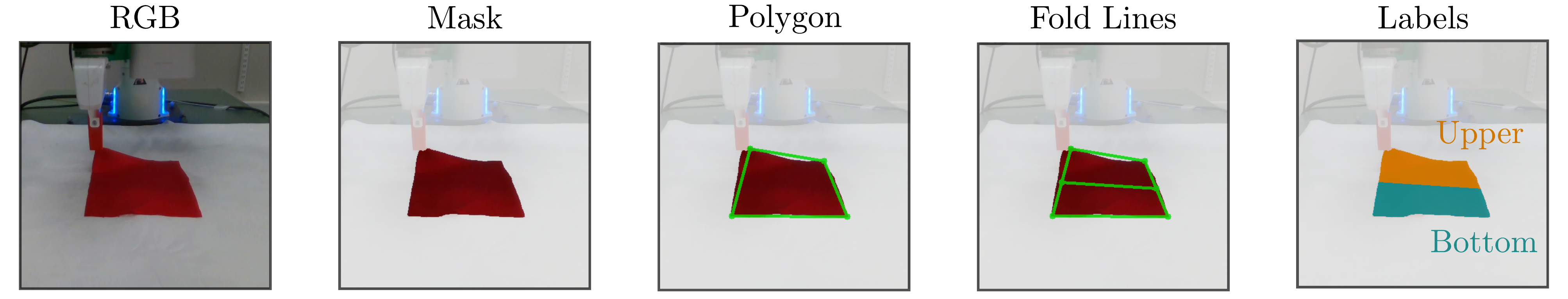}
    
 \vspace{-0.05in}
    \caption{\textbf{Cloth labeling:} Given the cloth mask at time $t=0$, we obtain fold lines based on geometric features following~\cite{miller2011parametrized}. These fold lines define the upper and bottom layers based on the pick and place locations.   
    }
    \label{fig:Ensamble}
    
  \vspace{-0.20in}
\end{figure*}
\label{sec:perception}

\section{Problem Formulation}

The problem we address is feedback-loop manipulation of cloth, focusing on optimizing the manipulation trajectory within a set of pick-and-place positions $x_{\text{pick}}, x_{\text{place}} \in \mathbb{R}^3$. These positions can be provided by a dedicated planner~\cite{lee2021learning, mo2022foldsformer, hoque2022visuospatial}. We consider the half-folding task proposed in~\cite{garcia2020benchmarking} as a representative evaluation task, where the goal is to fold a rectangular cloth in half. 
The task is performed with the assumption of quasi-static manipulation, which implies that the forces and torques acting on the cloth are in static equilibrium at each time-step. Under these conditions, we decouple the problem of feedback loop manipulation into cloth perception and trajectory optimization. \\
\noindent \textbf{Cloth perception:} The state of the cloth at time $t$ is described as a 3D point cloud $P_t$  representing the observable points of the cloth. The subscript $t$ is omitted when the specific time-step of the state is not necessary. 
To disambiguate different cloth configurations, we introduce semantic information in the point cloud, allowing us to cluster the points into two sets: $P^{U}$ and $P^B$,  where $P^U$ corresponds to the \textbf{U}pper layer points of the cloth, $P^B$ to the \textbf{B}ottom layer, and $P=P^U \cup P^B$. \\
\noindent \textbf{Trajectory Optimization:}  Consider the discrete-time dynamics of the observable points of a cloth ${P_{t+1} = f(P_t, x_t, a_t, \xi)}$, where $P_t=P_t^U \cup P_t^B$, 
$x_t$ is the 3D position of the robot end-effector (EE), $a_t$ is the robot action corresponding to a 3-DoF EE displacement, and $\xi$ are the cloth physical properties such as stiffness and elasticity.  The  folding trajectory from time $t=0$ to $t=T$ is defined as:
\begin{equation*}
\tau_{0:T} = 
\begin{bmatrix}
P_0 & P_1 & \dots & P_T \\
x_0 & x_1 & \dots & x_T \\
a_0 & a_1 & \dots & a_T
\end{bmatrix}.
\end{equation*}
The trajectory is optimized by finding a sequence of actions $a_{0:T}^*$ that minimizes an objective $\mathcal{J}$ factorized over per-timestep costs $c_t(P_t, x_t, a_t)$:

\begin{equation}
\begin{aligned}
a^*_{0:T} &= \underset{a_{0:T}}{\mathrm{arg\,min}} \ \mathcal{J}(\tau_{0:T}) = \underset{a_{0:T}}{\mathrm{arg\,min}} \ \sum_{t=0}^{T} c_t, \\
 \text{s.t.} & \quad  x_0 = x_{\text{pick}}  \\
& \quad x_T = x_{\text{place}}  \\
& \quad  P_{t+1} = f(P_t, x_t, a_t, \xi)   \\
& \quad  x_{t+1} = x_t + a_t 
\end{aligned}
\end{equation}
\noindent where $T$ is the control horizon and $\mathcal{J}(\tau_{0:T})$ denotes the objective value of the trajectory  $\tau_{0:T}$ . The control input is updated at every time step in a feedback-loop fashion using MPC with horizon $H$. Commonly, the cloth dynamics $f$ and the physical properties $\xi$ are not precisely known and vary across different objects~\cite{longhini2023elastic}. Thus, similarly to~\cite{longhini2022edo}, we jointly learn 1) an adaptation module $g_\psi$ to encode a recent history of observations into a latent representation $z_t$ of the physical properties and adapt to different objects, 2)  an approximate model  $f_\theta$ of the cloth dynamics conditioned on the latent representation. These models are respectively parameterized by $\theta$ and $\psi$. Our framework couples cloth perception with data-driven trajectory optimization for \textbf{Ada}ptive \textbf{Fold}ing, which we denote AdaFold.

\section{Cloth Perception}
\label{sec:perception}

The cloth perception consists of processing RGB-D observations into a semantically labelled point cloud $P = P^U \cup P^B $ representing the cloth.  {\new 
Specifically, our method involves two main steps: 1) extracting at time $t=0$ the masks representing the upper and bottom layers of the cloth using its geometric features, and 2) tracking these masks with pre-trained video tracking models.

Given an RGB-D image $I_0$ at time step $t=0$, we start by finding the segmentation mask of the full cloth $M_f$ from the RGB observation using an object segmentation module. In this work, we use Grounding-DINO \cite{liu2023grounding} and Segment Anything (SAM) \cite{kirillov2023segment} along with the prompt ``cloth". Given the binary mask of the cloth, we identify geometric features by fitting a parameterized polygon to the contour of the cloth as suggested in ~\cite{miller2011parametrized}. These geometric features, represented by the cloth corners for the specific case of rectangular cloths, allow the definition of task-specific folds by specifying a directed line segment in the plane that partitions the polygon into two parts, one to be folded over another~\cite{miller2012geometric}. To specifically implement half-folding, we project the pick-and-place positions $x_{\text{pick}}, x_{\text{place}}$ onto the image. These positions guide the selection of the fold segments that separate the cloth mask into upper  $M_u$ and bottom  $M_b$ halves, corresponding to the upper and bottom layers of the fold. Specifically, the upper mask includes $x_{\text{pick}}$ but not $x_{\text{place}}$, while the bottom mask includes $x_{\text{place}}$ but not $x_{\text{pick}}$.
Following this segmentation, we transform the masked depth observations into a point cloud $P_t$ using the camera's intrinsic matrix and label each point that belongs to the upper $P^U $and bottom $P^B $ layers according to the labels of the mask. An overview of this labeling process is detailed in Fig.~\ref{fig:Ensamble}. 

While extracting these features is straightforward from flattened configurations, the process becomes more challenging when the cloth is manipulated, as its shape undergoes deformation. We address this problem by tracking the full $M_f$ and upper $M_u$ masks of the cloth using video tracker XMEM~\cite{cheng2022xmem} at each time-step of the manipulation.  Given these two masks, the bottom mask is easily obtained as 
${M_b = M_f - M_u}$. 

}

\section{Trajectory Optimization}
To optimize the folding trajectory $\tau_{0:T}$ we use sampling-based MPC 
over a finite horizon $H$ using a learned model of the cloth dynamics.

\subsection{Learned Model}
To implement $f_\theta$ and $g_\psi$, we extend upon our previous work~\cite{longhini2022edo} with two main differences: 1) we substitute the Graph Neural Network (GNN)~\cite{scarselli2008graph} with PointNet++ \cite{qi2017pointnet++} to handle point clouds, 2) we obtain the latent representation $z_t$ in an online fashion by leveraging a recent history of past observations instead of a predefined exploratory action. As the forward model architecture is not the main contribution of this work, we refer the reader to \cite{longhini2022edo} for a more detailed explanation of this architecture.

We train the model on a dataset $\mathcal{D}$ of trajectories collected in simulation using a multi-step Mean Square Error (MSE) loss between the model predictions and the ground truth:
\begin{equation}
    \mathcal{L}(\theta, \psi) = \frac{1}{|\mathcal{D}|}  \sum_{\mathcal{D}} \left(  \frac{1}{M} \sum_{m=1}^{M} \frac{1}{2} \lVert \hat{P}_{t+m}- P_{t+m} \rVert^2  \right),
    \label{eq_loss}
\end{equation}
where $ \hat{P}_{t+m}=f_\theta(\hat{P}_{t+m-1}, a_t, z_t) $, $z_t = g_\psi(\tau_{t-K:t-1})$ {\new, $K$ is the length of the recent history use to encode the latent representation $z_t$ and  $M$ represents the number of steps in the future to be predicted.} We use the MSE over the Chamfer loss as we empirically observed more stable training without compromising accuracy. Finally, we assume that ${\hat{P}^B_{t+1} ={P}^B_{t}}$ due to the friction between the cloth and the table, preventing the bottom half from slipping during manipulation. Thus, it is sufficient for the model to predict only $\hat{P}^U_{t+1}$.

\label{sec:perception}

\subsection{Optimization Process}
\label{sec:optimization}

A scheme of Adafold is proposed in Algorithm~\ref{alg:Adafold}. Next, we outline the algorithm, detail the cost function design, describe a constrained sampling strategy to reduce the action search space, and introduce a heuristic to ensure candidate trajectories end at the place position. \\
\noindent\textbf{Algorithm:} At each time $t$, the perception modules first process the cloth representation $P_t$ (line $2$). Then, $N$ candidate open-loop control sequences $a_{t:t+H}$ are sampled from a multivariate Gaussian distribution (line $4$). The cost of each control sequence $ \mathcal{J}^n(\tau_{t:t+H}) $ is computed using the trajectory $\tau_{t:t+H}$ predicted by the model rollout (line $6-10$). The optimal control input is finally updated using the MPPI algorithm \cite{williams2017information}, which weights the sampled candidate control sequences following (line $12$):
\begin{equation}
a^*_h = \frac{1}{\sum_{n=1}^{N} \exp\left(-\frac{1}{\lambda} \mathcal{J}^n\right)} \sum_{n=1}^{N} \exp\left(-\frac{1}{\lambda} \mathcal{J}^n\right) a^n_h,
\label{eq:MPPI_update}
\end{equation}
for $h=t, .., t+H$ where $\mathcal{J}^n$ is the cost of the $n$-th trajectory, $\lambda$ is a temperature parameter that controls exploration, and $a^n_t$ is the control input of the $n$-th trajectory at time step $t$. Finally, only the first updated control input $a^*_t$ is executed, while the un-executed portion of the optimized trajectory is used to warm-start the optimization at time $t+1$ (lines $13-14$).

\begin{algorithm}[t]
\SetAlgoLined
\KwResult{Optimized folding actions $a^*_{0:T}$.}
\KwIn{Pick and place positions $\{x_{\text{pick}}, x_{\text{place}}\}$, Learned models $f_\theta$ and $g_\psi$, Horizon $H$, Number of action candidates $N$, Control hyper-parameters $\lambda, w_1, w_2$, Initial control sequence $a_{0:H}$, Control variance: $\Sigma$}
\For{$t \gets 0$ \KwTo $T$}{
    $P_t = P^U_t \cup P^B_t \gets \text{Cloth Perception}(\{I^1_t, I^2_t\})$\\
    \For{$n \gets 1$ \KwTo $N$}{
       $a^n_{t:t+H} \gets \mathcal{N}(a_{t:t+H}, \Sigma)$ \Comment{Constrained}\\
   
        $z_t \gets g_\psi(\tau_{t-K:t-1}) $\\        
        \For{$h \gets t$ \KwTo $t+H$}{
            $\text{Compute} \quad \tau_{h:h+1} \quad  \text{unrolling } f_\theta $  \\ 
            $\mathcal{J}^n(\tau_{h:h+1}) \gets  w_1 c_1+ w_2 c_2$ \Comment{Eq. (~\ref{eq:J})}\\
        }
        $\mathcal{J}^n(\tau_{t:t+H}) \gets \sum_{h=t}^{t+H} \mathcal{J}^n(\tau_{h:h+1})$\\

    }
  $a^*_{t:t+H} \gets \text{MPPI}(\{\mathcal{J}^n(\tau_{t:t+H})\}_{n=1}^N, \lambda)$ \Comment{Eq. (~\ref{eq:MPPI_update})}\\

      $\text{Execute} \quad a^*_t$\\
      $\text{Warm-start control sequence }  a   \text{ with }  a^*$

}
\caption{AdaFold}
\label{alg:Adafold}
\end{algorithm}

\noindent\textbf{Cost function:} The cost function is designed to select actions that maximize the alignment between the two halves of the cloth. This alignment is quantitatively assessed using the Intersection over Union (IoU) metric, calculated based on the {\new ratio of the} areas occupied by each half $P^U$ and $P^B$ when projected to the folding plane. 
We additionally incorporate in the cost function a term discouraging actions resulting in large displacements of the cloth. Specifically, the cost function is defined as a weighted sum of two cost terms:
\begin{equation}
\begin{aligned}
\mathcal{J}(\tau_{t:t+H})= w_1 c_1(\tau_{t:t+H}) + w_2 c_2(\tau_{t:t+H}),
\end{aligned}
\label{eq:J}
\end{equation}
where the weight vectors $w_1, w_2$ allow us to weigh differently the desired behaviors.
The first term $c_1$ evaluates the progress towards the alignment of the two halves as:
\begin{equation}
  {c_1(\tau_{t:t+H}) = \sum_{j=1}^{H } \beta^{H-j} \text{IoU}(\hat{P}^U_{t+j}, P^B_{t+j})},  
\end{equation}
 where $\beta$ is a weighting factor within the interval $(0, 1)$. In particular, $\beta^{H-j}$ has the role of progressively increasing the importance of future cloth alignments as its value increases while $j$ increases. This design choice prevents the robot from greedily selecting actions that lead to aligning the halves as fast as possible, compromising the future quality of the fold. 
The second term $c_2$, instead,  acts as a binary flag that assesses whether the action $a_t$ leads the gripper's subsequent position $x_{t+1}$ outside a predefined convex hull defined by the initial shape of the flattened cloth. In practice,  $c_2$ is set to $1$ if the gripper exits the specified convex hull, while it is set to $0$ otherwise. Intuitively, if the gripper pulls the corner towards a region beyond the initial cloth shape, the action results in a larger movement of the entire cloth. \\ 
\noindent\textbf{Constrained random sampling:} 
To reduce the action search space, we constrain the sampling of the random candidate control sequence at time $t$ by sampling directions $\mathbf{d}$ similar to the pick-and-place direction $\mathbf{d}_{pp}$. Specifically, we impose the cosine similarity ($\text{CS}$) to satisfy $\text{CS}(\mathbf{d}, \mathbf{d}_{pp}) \geq 0$, {\new which enforces the actions to always move towards the place location}. The actions are then defined as $a = (\mathbf{d} / \lVert \mathbf{d}\lVert ) v$ where $v$ is a fixed value specifying the action norm. \\
\textbf{Place reaching:} If the termination of the sampled trajectory is not sufficiently close to the place position, we use a heuristic to extend the path. This heuristic continues the trajectory along a linear path towards $x_{\text{place}}$ Fig. \ref{fig:method_all} shows an example of candidate trajectories (light blue) terminating in the place position. While extending the trajectory to the place position is a reasonable assumption since $x_{\text{place}}$ is known, it could be removed by incorporating an attractor towards the place position into the cost function, with the disadvantage of an additional hyperparameter to tune.\\

\section{Implementation Details}

\subsection{Half folding task}

\noindent \textbf{Overview:} We consider the half folding task proposed in~~\cite{garcia2020benchmarking}. In this task, the agent controls a square or rectangular cloth placed approximately in the center of the environment at the beginning of each episode. We assume the agent starts from a
grasped state, while the objective is to fold the cloth in half by aligning the corners. \\ 
\noindent \textbf{Success Criteria:} We evaluate the success of the folding execution by computing the 2D intersection over union (IoU) between the two halves of the cloth as described in Section~\ref{sec:optimization}. As the bottom half of the cloth will be occluded in later stages of the manipulation, we use as a reference the points observed at time $t=0$.

\subsection{Simulation}

We implemented the half-folding task in PyBullet~\cite{benelot2018}. To collect a training dataset of state-action trajectories,  we selected a square cloth of length $20$cm  with elastic and stiffness parameters of $40$ and $60$, respectively. We include variations of these two parameters in the interval $[20, 100]$ in the test phase. 
We considered pick and place positions in the top-left and bottom-left corners, as selecting specific pick and place locations for data collection has shown to be beneficial to learn transition models for folding ~\cite{hoque2022visuospatial}. We then generated $1000$ pick-and-place trajectories $\tau$ with $T=12$ using the constrained random sampling described in Sec.~\ref{sec:optimization}, resulting in  $\sim 15$k training data points. We further augment the point clouds during training by applying random scaling, rotations, translations, and adding Gaussian noise.

\begin{table}[t!]
\centering
\caption{Properties of the real-world cloths, measured as specified in~~\cite{garcia2024standard}. CT, Stiff. and Elast. stand for respectively construction technique, stiffness and elasticity.}
\begin{tabular}{lccccc}
\toprule
\textbf{Cloth} & \textbf{Size} $[\text{cm}]^2$  & \textbf{Material}  & \textbf{CT}  & \textbf{Stiff.}  & \textbf{Elast.} \\
\midrule
1 &  $20 \times 30 $ & Polyester & Woven-Plain & 0.41 &  0.03\\
2  & $17 \times 25 $ & Polyester & Woven-Plain & 0.56 & 0.40   \\
3 & $17 \times 25 $  & Cotton  &  Knitted & 0.69 &  0.36\\
4 & $17 \times 25 $  & Wool & Knitted & 0.71  & 0.68 \\
5 & $17 \times 25 $  &  Cotton & Woven-Twill & 0.78 & 0.08  \\
6 & $17 \times 25 $  & Cotton &  Woven-Plain & 0.74 & 0.06\\
\new{7} & \new{$28 \times 20 $ } &\new{Cotton} &  \new{Woven-Plain} &\new{0.87} &\new{0.22}\\
\bottomrule
\end{tabular}
\label{tab:properties}

  \vspace{-0.15in}
\end{table}

\begin{figure}[t!]
    \centering
    \includegraphics[width=0.3\textwidth]{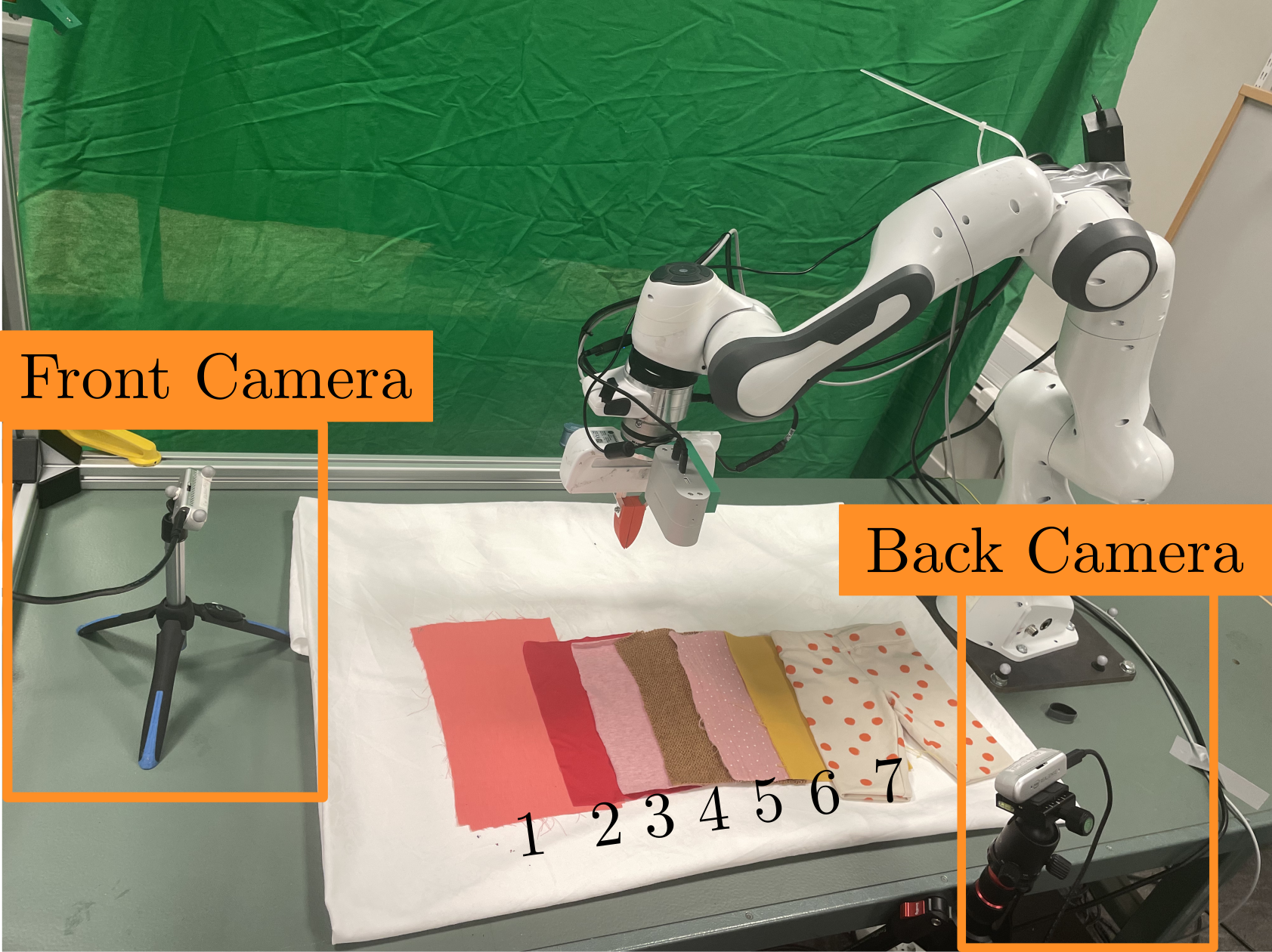}
    \caption{  Visualization of the real-world set-up: two Realsense D435 cameras capturing different views of the scene, and the dataset composed of {\new $7$} cloths. }
    \label{fig:set_up}

  \vspace{-0.15in}
\end{figure}

\subsection{Real-world} 
The real-world set-up and the samples used for the experiments are visualized in Fig.~\ref{fig:set_up}.\\
\noindent \textbf{Dataset:} We utilized six rectangular cloths with different material, construction, and physical properties~\cite{longhini2021textile}, none of them matching the one used in the simulation, {\new and one pair of pants} (see Fig.~\ref{fig:set_up}). Specifically, cloths {\new $2-6$} share the same shape but differ in physical properties. {\new Cloth  $1$ retains the rectangular shape, but it is bigger than the others. Cloth  $7$ differs in shape and size. } 
We report the cloth properties in Table~\ref{tab:properties}. 
\\
\noindent \textbf{Perception:} We collected observations from two calibrated cameras. {\new To find the geometric features from the mask of the full cloth, 
we fit a polygon with respectively $4$ and $7$ vertices to the rectangular and pants objects. We then constructed the point cloud based on the camera intrinsics and extrinsic parameters.} 
We further filtered the point cloud to remove outliers and voxelize it with a voxel size of $0.8$cm. {\new To track the masks, we used an open-source pre-trained video tracker available at ~\cite{cheng2022xmem}. } \\

\subsection{Modeling and Planning}
\noindent \textbf{Modeling:} We used the official implementation of the segmentation and classification branches of PointNet++~\cite{qi2017pointnet++} for the forward and adaptation network architectures, respectively. 
We considered $K = 3$ past observations and $M=3$ future predictions for the multi-step MSE loss,  while the dimension of the latent variable $z$ was set to $32$. We trained the models for $400$ epochs, with a learning rate of $0.001$ and a batch size equal to $32$. \\
\noindent \textbf{Planning:} We set the horizon to $H = 12$ and considered $N=100$ candidate trajectories drawn from a multivariate Gaussian distribution initialized at time $t=0$ with zero mean and diagonal covariance matrix set to $0.01$. At the time $t>0$, we kept the covariance of the distribution fixed while the mean was updated using Eq.\ref{eq:MPPI_update} with temperature parameter $\lambda = 0.01$. We set the remaining planning hyperparameters to $w_1=1, w_2=0.03$, $\beta = 0.5$, and $v=0.03$.

\begin{table}[t]
\centering
\caption{Comparative folding results showcasing the final IoU. The folding is executed 20 times for each combination of cloth and methods, where $N=10$.}
\begin{tabular}{lcc}
\toprule
\textbf{Method} & {1 Cloth} $\uparrow$ & {N Cloths} $\uparrow$\\
\midrule
Random &  $0.39 \pm 0.14 $ & $0.40 \pm 0.17$ \\
Triangular  & $0.41 \pm 0.00$ & $0.41 \pm 0.02$  \\
DDPG-Critic & $0.49 \pm 0.09$  & $0.48 \pm 0.09$ \\
\midrule
Adafold-OL  & $0.53 \pm 0.17 $  &  $0.48 \pm 0.16$ \\
\new{Adafold-NL}  & \new{$0.57 \pm 0.08 $}  &  \new{$0.57 \pm 0.09$} \\
\rowcolor{lavender}
AdaFold  & $\mathbf{0.83 \pm 0.09}$  & $\mathbf{0.78 \pm 0.11}$  \\
\bottomrule
\end{tabular}
\label{tab:planning}

\end{table}

\begin{figure}[t!]
    \centering
    \includegraphics[width=0.48\textwidth]{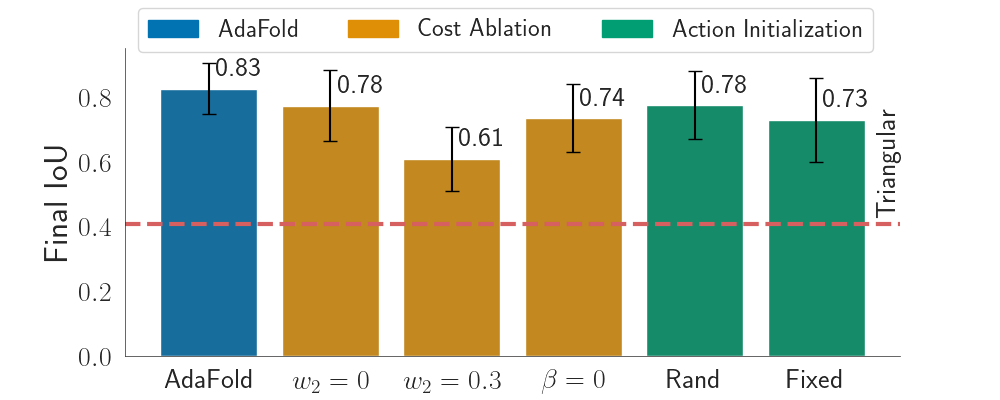}
    \caption{Cost and action initialization ablation.
    The reference performance of the Triangular trajectory is shown as the red horizontal dashed line. The fold is executed 20 times for each ablation.}
    \label{fig:sub_ablation_cost}

  \vspace{-0.15in}
\end{figure}

\section{Experimental Results}

Our experiments aim to study the effectiveness of AdaFold to optimize folding trajectories via feedback-loop manipulation. In particular, we investigate to what degree our proposed approach: 1) improves the folding outcome, 2) generalizes to cloth with variations in physical properties, and 3) improves the particle-based representation of the cloth through semantic labels. 

\subsection{Baselines}
We compare AdaFold against {\new five} different baselines. The first baseline, {\new representing most of open loop methods}, is a fixed triangular trajectory (Triangular), which was selected as the best performing in simulation when compared to a linear trajectory. The second is a random baseline (Random), which, at each time step, randomly selects an action among all the candidate actions. The third {\new and fourth baselines are ablations of AdaFold, one considering an open-loop version of AdaFold (AdaFold-OL) that optimizes the trajectory once at the beginning of the folding, and one that uses a point cloud without labels (AdaFold-NL)}. The final baseline we compared to was the model-free learning method DDPG \cite{lillicrap2015continuous}. For a fair comparison, we trained DDPG offline on the same dataset as AdaFold to learn a state-action value function. At test time, we used the learned value function to select the best action among the candidates (DDPG-Critic). The reward function and the model architecture for this baseline correspond to the one chosen for our approach, where the critic architecture and hyperparameters match the ones of AdaFold. As the output of the critic has the same dimension as the number of input points, to extract the value from the output of the segmentation branch, we select the first point of the network output as similarly done in \cite{wang2023one}.

\subsection{Folding Optimization Results}

The goal of this section is two-fold: assessing the relevance of optimizing the folding trajectory and showcasing the benefits of feedback-loop optimization. We conducted the experiment in simulation, and we compared our method against the {\new five} baselines presented in the previous section.
We measured the folding success using the IoU metric. We evaluated all the methods under two conditions: one where the folding task utilized the same cloth as in training and another where $10$ cloth parameters were randomly selected to introduce variations not observed during training. 
We repeated the folding 20 times for each combination of cloth parameters and method.

\begin{figure}[t!]
    \centering
    \includegraphics[width=0.48\textwidth]{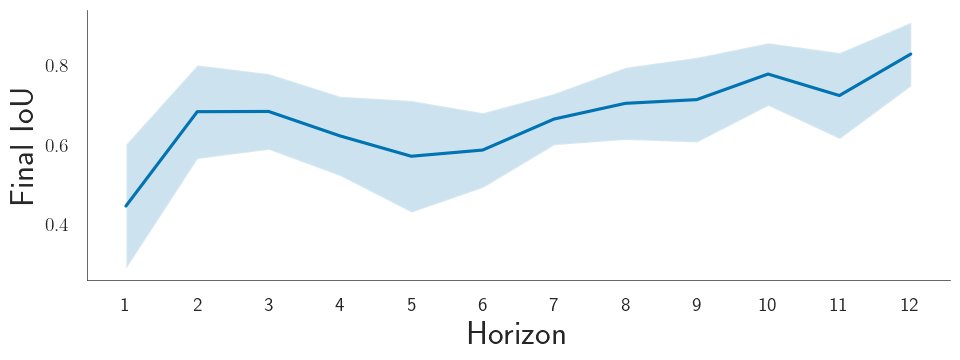}
    \caption{Evaluation of different planning horizons, where the task length is $T=12$. For each horizon, the fold is executed 20 times.}
    \label{fig:sub_horizon_cost}

\end{figure}

Table \ref{tab:planning} shows the folding results. It can be observed that methods that do not optimize the folding trajectory, specifically the Random and Fixed baselines, yielded the lowest performance, thereby underscoring the advantage of trajectory optimization. The Random baselines had, on average, a similar performance to the Triangular baseline, as our constrained sampling approach favours some alignment at the end of the execution by completing the random trajectory till the place position. While the model free-baselines (DDPG-Critic) improved the fold compared to the Triangular baseline, the small amount of offline data proved to be a challenging scenario for learning a good value function for folding. A similar result can be observed for {\new AdaFold's ablations (AdaFold-OL, AdaFold-NL)}, which improved the fold compared to the fixed trajectory but obtained, on average, a worse result than the feedback-loop variant.  AdaFold outperformed all the baselines and achieved the best folds, showcasing the benefits of feedback-loop optimization. 
Finally, the small difference between the results over $1$ or $N$ cloths of the Triangular baseline suggests that the variation in cloth behaviours that can be simulated by PyBullet is not large, as studied in~\cite{blanco2023benchmarking}. Thus,  while AdaFold folding success for $N$ cloths still outperformed the Triangular trajectory, we defer the evaluation of AdaFold's generalization to cloths with varying physical properties to the real-world experiments.

\begin{table}[t]
\centering
\caption{Final IoU evaluated on real-world Cloths {\new (C.)} $2-6$. The folding is repeated 5 times for each combination of cloth and method. {\new Best viewed with zoom}. }
\resizebox{0.49\textwidth}{!}{
\begin{tabular}{lccccc}

\toprule
\new{\textbf{C.}} & \textbf{Triangular $\uparrow$} & \new{\textbf{DDPG-C $\uparrow$}} & \new{\textbf{AdaFold-OL$\uparrow$}} & \new{\textbf{AdaFold-NL $\uparrow$}} & \textbf{AdaFold $\uparrow$} \\
\midrule
2 & \new{$0.63 \pm 0.03$} & \new{$0.39 \pm 0.10$} & \new{$0.65 \pm 0.18$} & \new{$0.51 \pm 0.07$} & \new{$\mathbf{0.72 \pm 0.07}$} \\
3 & \new{$0.53 \pm 0.04$} & \new{$0.33 \pm 0.06$} & \new{$0.61 \pm 0.08$} & \new{$0.58 \pm 0.04$} & \new{$\mathbf{0.65 \pm 0.06}$} \\
4 & \new{$0.61 \pm 0.01$} & \new{$0.35 \pm 0.18$} & \new{$0.70 \pm 0.06$} & \new{$0.63 \pm 0.02$} & \new{$\mathbf{0.74 \pm 0.04}$} \\
5 & \new{$0.72 \pm 0.09$} & \new{$0.37 \pm 0.02$} & \new{$0.52 \pm 0.12$} & \new{$0.59 \pm 0.03$} & \new{$\mathbf{0.74 \pm 0.06}$} \\
6 & \new{$0.75 \pm 0.11$} & \new{$0.45 \pm 0.16$} & \new{$0.77 \pm 0.09$} & \new{$0.64 \pm 0.03$} & \new{$\mathbf{0.81 \pm 0.05}$} \\
\bottomrule
\end{tabular}
}
\label{tab:rw_planning2}

\end{table}

\begin{table}[t]
\centering
\caption{Evaluation of AdaFold's generalization to variations of cloth initial positions and size. The evaluation metric is the IoU. The folding is repeated 1 time for each random position, and 5 times for cloth $1$ for both evaluated methods. }
\resizebox{0.49\textwidth}{!}{
\begin{tabular}{lcccc}
\toprule
\textbf{Method} & \textbf{10 Poses} $\uparrow$ & \textbf{Cloth 1} $\uparrow$ & \new{\textbf{Leg}} $\uparrow$ & \new{\textbf{Pants}} $\uparrow$\\
\midrule
Triangular  & \new{$0.63 \pm 0.03$} & \new{$ 0.64  \pm 0.03$} &  \new{$ 0.58 \pm 0.05$} & \new{$ 0.64\pm 0.03$}  \\
\rowcolor{lavender}
AdaFold  & \new{$\mathbf{0.72 \pm 0.08}$}  & \new{$ \mathbf{0.76  \pm 0.04 } $}  & \new{ $\mathbf{0.64 \pm 0.04}$} & \new{$ \mathbf{0.68 \pm 0.05}$} \\
\bottomrule
\end{tabular}
}
\label{tab:planning_more}

\end{table}

\subsection{Ablation Study}

In this section, we examine the impact of various design choices within our methodology. First, we explored the influence of individual components of the cost function, assessing how the folding performance is affected by setting the hyperparameters $\beta, w_2$ to zero or by increasing $w_2$ tenfold. Additionally, we evaluated the impact different initializations of the multivariate Gaussian distribution have on the resulting candidate actions. The initialization approaches we examined, in addition to the zero initialization, were: random and the predefined triangular trajectory. Fig.~\ref{fig:sub_ablation_cost} presents the results of the ablation. It can be observed that removing any of the components decreased the overall quality of the fold, as well as increasing the weight of $c_2$. Moreover, the initialization of the means of the multivariate Gaussian distribution significantly influenced the performance of our approach due to the sampling-based nature of the MPPI algorithm. Among the tested strategies, zero mean initialization demonstrated superior performance over random and fixed initialization, facilitating more efficient exploration of the control space by the MPPI algorithm. Yet, the performance of all the ablations consistently outperformed the Triangular baseline, suggesting that minimal tuning of the cost function can still enhance the performance relative to the baseline.    
Finally, we evaluated different lengths of the planning horizon $H$.  The results presented in Fig.~\ref{fig:sub_horizon_cost} show that increasing the planning horizon led to better folds, suggesting that a shorter horizon leads to myopic behaviours.

\subsection{Real World Experiments}

This evaluation investigates to what extent AdaFold generalizes to different cloth variations including physical properties, position, size and shape.

{\new \noindent \textbf{Physical properties:}} First, we evaluated variations in physical properties by assessing whether AdaFold can improve the fold quality compared to the Triangular trajectory for cloths $2-6$. 
We evaluated the folding outcome with the IoU metric. We present the results in Table \ref{tab:rw_planning2}. The variance that the Triangular trajectory exhibits highlights that identical folding trajectories can yield different outcomes based on the cloth's properties, where stiffer cloths (e.g. samples $5-6$) achieved on average a better fold with respect to less stiff cloths. Conversely, AdaFold consistently produced satisfactory folds across different samples, improving the fold for samples where the fixed trajectory fell short. Figure \ref{fig:intro_example} provides a qualitative comparison of the fixed and optimized trajectories for samples $2$ and $6$.

{\new \noindent \textbf{Position and size:}} We extended this evaluation by testing AdaFold on variations in the initial positions and shapes of the cloth. For the former variation, we selected cloth $2$ and randomly rotated its starting position on the table $10$ times, with rotations ranging between $\pm 45^{\circ} $ degrees. For the latter variation, we selected cloth $1$, which is larger than clothes $2-6$. We compared AdaFold against the Triangular baseline.  The results, detailed in Table \ref{tab:planning_more} {\new under the columns ``10 Poses" and ``Cloth 1"}, show AdaFold outperforming the Triangular trajectory in both scenarios. 

{\new \noindent \textbf{Shape:} We finally tested AdaFold's ability to handle different shapes and more complex folding using a pair of pants.  We designed two different scenarios: 1) folding one leg in half, where the overall shape differs from a rectangle, and 2) folding in half the pants with both legs already folded, where the shape is still rectangular but the cloth is already folded in multiple layers.  The results, detailed in Table \ref{tab:planning_more} under the columns ``Leg" and ``Pants", show AdaFold outperforming the Triangular baseline, confirming its versatility with various shapes and pre-folded configurations.} 

The overall outcome confirms the adaptability of our method, underscoring the benefit of using particle-based representations along feedback-loop manipulation to adapt to different object variations. 

\begin{figure}[t]
    \centering
    \includegraphics[width=0.42\textwidth]{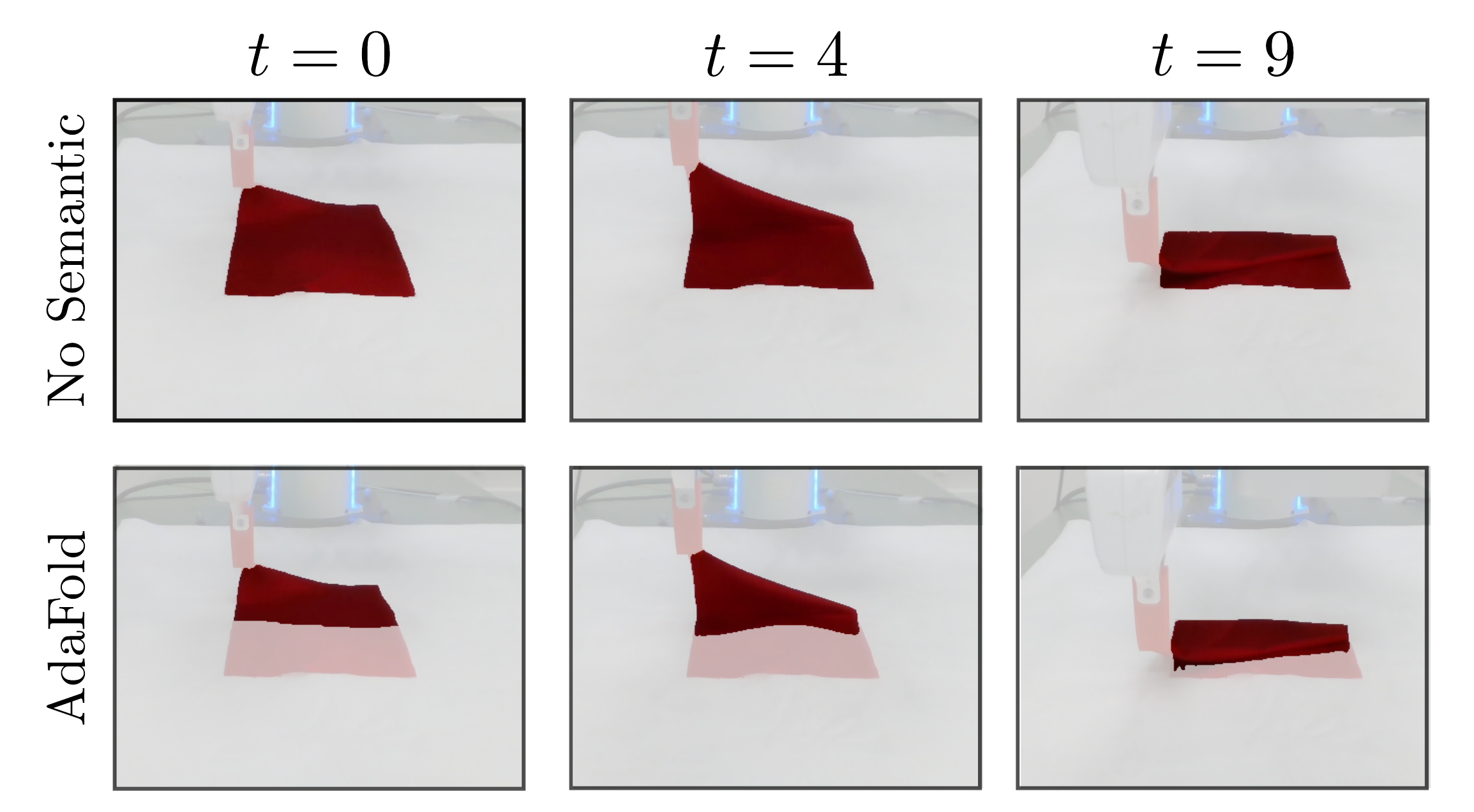}
    \caption{Visualization of: (top) the mask with no semantic tracked over time, (bottom) AdaFold upper mask tracked over time.}
    \label{fig:segmentation_results}
    
\end{figure}

\begin{table}[t]
\centering
\caption{Evaluation of the MAE of the IoU estimated with different point cloud representations.}
\begin{tabular}{lccc}
\toprule
\textbf{Camera(s)} & \textbf{No Semantic} $\downarrow$  & \new{\textbf{VLM}} $\downarrow$   & \new{\textbf{AdaFold}} $\downarrow$  \\
\midrule
Front &  \new{$0.39 \pm 0.10 $} &\new{$0.39 \pm 0.11$} & \new{$0.09 \pm 0.02 $} \\
Back  &\new{ $0.15 \pm 0.07 $} & \new{$0.15 \pm 0.04 $} & \new{$0.09 \pm 0.05$}  \\
Both & \new{$0.21 \pm 0.09  $}  &  \new{$0.23 \pm 0.08 $} & \new{$\mathbf{0.04 \pm 0.05}$}   \\
\bottomrule
\end{tabular}
\label{tab:perception}

\end{table}

\subsection{Semantic Cloth Representation}
Finally, we assess whether our proposed perception module improves the representation of the cloth. Unlike previous sections, the IoU is used as a quantitative description of the cloth state instead of a performance metric. As an evaluation dataset, we used a subset of the real-world trajectories recorded for the previous evaluations. We manually annotated the masks belonging to the upper or bottom half of the cloth to extract $P^U$ and $P^B$ to compute the ground truth IoU. To compare different state representations, we used the mean absolute error (MAE) computed between the estimated IoU from a specific cloth representation and the ground truth. Specifically, we compared the representations with semantic descriptors obtained through our {\new perception module (AdaFold)}, against a baseline representation lacking semantic descriptors (No Semantic), and a segmentation method derived from a greedy selection of the highest confidence mask from the {\new same VLM used for object segmentation at $t=0$ with prompts ``upper half cloth", ``bottom half cloth" (VLM)}. We further integrated the comparison between one and two camera points of view. 
As shown in Table \ref{tab:perception}, our approach always achieved the lowest MAE, with the best performance provided by leveraging the observations from \textit{both} cameras. The representation with no semantic information obtained the worst results, confirming the challenges of distinguishing different folded states only relying on the observed points of the cloth. 
{\new 
Integrating semantic information from the VLM did not yield good results, as the model always suggested the full cloth as a mask instead of the desired cloth regions. Trying to tune the prompts did not provide any improvement. 
In contrast, our method successfully segmented the desired cloth region.  Fig. \ref{fig:segmentation_results} shows qualitative results of the obtained masks.} 
These results confirm that augmenting point clouds with semantic descriptors provides a better representation of different folded configurations. 


\section{Conclusion and Future Work}

AdaFold framework leverages model-based feedback-loop manipulation to optimize cloth folding trajectories. It integrates semantic descriptors extracted from {\new geometric features} into point cloud representations. The experiments validated the hypothesis that AdaFold adapts folding trajectories to variations in physical properties, positions, and sizes of cloths. It was further showcased the potential of coupling a strong perception module with data-driven optimization strategies to perform feedback-loop manipulation. We plan to extend these results to different goal and reward configurations and a broader spectrum of manipulation tasks. We will investigate the integration of model-based and model-free learning approaches to address the computational demands of model-based planning. 


\bibliography{references}

\begin{thebibliography}{10}
\providecommand{\url}[1]{#1}
\csname url@rmstyle\endcsname
\providecommand{\newblock}{\relax}
\providecommand{\bibinfo}[2]{#2}
\providecommand\BIBentrySTDinterwordspacing{\spaceskip=0pt\relax}
\providecommand\BIBentryALTinterwordstretchfactor{4}
\providecommand\BIBentryALTinterwordspacing{\spaceskip=\fontdimen2\font plus
\BIBentryALTinterwordstretchfactor\fontdimen3\font minus \fontdimen4\font\relax}
\providecommand\BIBforeignlanguage[2]{{%
\expandafter\ifx\csname l@#1\endcsname\relax
\typeout{** WARNING: IEEEtran.bst: No hyphenation pattern has been}%
\typeout{** loaded for the language `#1'. Using the pattern for}%
\typeout{** the default language instead.}%
\else
\language=\csname l@#1\endcsname
\fi
#2}}

\bibitem{kroemer2021review}
O.~Kroemer, S.~Niekum, and G.~Konidaris, ``A review of robot learning for manipulation: Challenges, representations, and algorithms,'' \emph{JMLR}, vol.~22, no.~1, pp. 1395--1476, 2021.

\bibitem{zhu2022challenges}
J.~Zhu, A.~Cherubini, C.~Dune, D.~Navarro-Alarcon, F.~Alambeigi, D.~Berenson, F.~Ficuciello, K.~Harada, J.~Kober, X.~Li, \emph{et~al.}, ``Challenges and outlook in robotic manipulation of deformable objects,'' \emph{Robotics \& Automation Magazine}, vol.~29, no.~3, pp. 67--77, 2022.

\bibitem{longhini2024unfolding}
A.~Longhini, Y.~Wang, I.~Garcia-Camacho, D.~Blanco-Mulero, M.~Moletta, M.~Welle, G.~Aleny{\`a}, H.~Yin, Z.~Erickson, D.~Held, \emph{et~al.}, ``Unfolding the literature: A review of robotic cloth manipulation,'' \emph{arXiv preprint arXiv:2407.01361}, 2024.

\bibitem{lin2022learning}
X.~Lin, Y.~Wang, Z.~Huang, and D.~Held, ``Learning visible connectivity dynamics for cloth smoothing,'' in \emph{CoRL}, 2022, pp. 256--266.

\bibitem{ma2021learning}
X.~Ma, D.~Hsu, and W.~S. Lee, ``Learning latent graph dynamics for deformable object manipulation,'' \emph{arXiv:2104.12149}, vol.~2, 2021.

\bibitem{miller2012geometric}
S.~Miller, J.~Van Den~Berg, M.~Fritz, T.~Darrell, K.~Goldberg, and P.~Abbeel, ``A geometric approach to robotic laundry folding,'' \emph{IJRR}, vol. 31(2), pp. 249--267, 2012.

\bibitem{miller2011parametrized}
S.~Miller, M.~Fritz, T.~Darrell, and P.~Abbeel, ``Parametrized shape models for clothing,'' in \emph{IEEE ICRA}, 2011, pp. 4861--4868.

\bibitem{camacho2013model}
E.~F. Camacho and C.~B. Alba, \emph{Model predictive control}.\hskip 1em plus 0.5em minus 0.4em\relax Springer science \& business media, 2013.

\bibitem{cheng2022xmem}
H.~K. Cheng and A.~G. Schwing, ``Xmem: Long-term video object segmentation with an atkinson-shiffrin memory model,'' in \emph{ECCV}.\hskip 1em plus 0.5em minus 0.4em\relax Springer, 2022, pp. 640--658.

\bibitem{garcia2020benchmarking}
I.~Garcia-Camacho, M.~Lippi, M.~C. Welle, H.~Yin, R.~Antonova, A.~Varava, J.~Borras, C.~Torras, A.~Marino, G.~Alenya, \emph{et~al.}, ``Benchmarking bimanual cloth manipulation,'' \emph{RA-L}, vol.~5, no.~2, pp. 1111--1118, 2020.

\bibitem{moletta2022representing}
M.~Moletta, M.~C. Welle, A.~Kravchenko, A.~Varava, and D.~Kragic, ``Representing clothing items for robotics tasks,'' \emph{KTH Royal Institute of Technology}, 2022.

\bibitem{yan2021learning}
W.~Yan, A.~Vangipuram, P.~Abbeel, and L.~Pinto, ``Learning predictive representations for deformable objects using contrastive estimation,'' in \emph{CoRL}.\hskip 1em plus 0.5em minus 0.4em\relax PMLR, 2021, pp. 564--574.

\bibitem{hoque2022visuospatial}
R.~Hoque, D.~Seita, A.~Balakrishna, A.~Ganapathi, A.~K. Tanwani, N.~Jamali, K.~Yamane, S.~Iba, and K.~Goldberg, ``Visuospatial foresight for physical sequential fabric manipulation,'' \emph{Autonomous Robots}, pp. 1--25, 2022.

\bibitem{huang2023self}
Z.~Huang, X.~Lin, and D.~Held, ``Self-supervised cloth reconstruction via action-conditioned cloth tracking,'' \emph{arXiv:2302.09502}, 2023.

\bibitem{popovic2011grasping}
P.~Mila, G.~Kootstra, J.~A. J{\o}rgensen, D.~Kragic, and N.~Kr{\"u}ger, ``Grasping unknown objects using an early cognitive vision system for general scene understanding,'' in \emph{IEEE/RSJ IROS}, 2011, pp. 987--994.

\bibitem{wang2023one}
Y.~Wang, Z.~Sun, Z.~Erickson, and D.~Held, ``One policy to dress them all: Learning to dress people with diverse poses and garments,'' \emph{arXiv:2306.12372}, 2023.

\bibitem{willimon2011model}
B.~Willimon, S.~Birchfield, and I.~Walker, ``Model for unfolding laundry using interactive perception,'' in \emph{IEEE/RSJ IROS}, 2011, pp. 4871--4876.

\bibitem{qiu2023robotic}
Y.~Qiu, J.~Zhu, C.~Della~Santina, M.~Gienger, and J.~Kober, ``Robotic fabric flattening with wrinkle direction detection,'' \emph{arXiv:2303.04909}, 2023.

\bibitem{seita2020deep}
D.~Seita, A.~Ganapathi, R.~Hoque, M.~Hwang, E.~Cen, A.~K. Tanwani, A.~Balakrishna, B.~Thananjeyan, J.~Ichnowski, N.~Jamali, \emph{et~al.}, ``Deep imitation learning of sequential fabric smoothing from an algorithmic supervisor,'' in \emph{IEEE/RSJ IROS}, 2020, pp. 9651--9658.

\bibitem{wu2019learning}
Y.~Wu, W.~Yan, T.~Kurutach, L.~Pinto, and P.~Abbeel, ``Learning to manipulate deformable objects without demonstrations,'' \emph{arXiv:1910.13439}, 2019.

\bibitem{mo2023learning}
K.~Mo, Y.~Deng, C.~Xia, and X.~Wang, ``Learning language-conditioned deformable object manipulation with graph dynamics,'' \emph{arXiv:2303.01310}, 2023.

\bibitem{petrik2019feedback}
V.~Petr{\'\i}k and V.~Kyrki, ``Feedback-based fabric strip folding,'' in \emph{IEEE/RSJ IROS}, 2019, pp. 773--778.

\bibitem{hietala2022learning}
J.~Hietala, D.~Blanco-Mulero, G.~Alcan, and V.~Kyrki, ``Learning visual feedback control for dynamic cloth folding,'' in \emph{IEEE/RSJ IROS}, 2022, pp. 1455--1462.

\bibitem{duan2016benchmarking}
Y.~Duan, X.~Chen, R.~Houthooft, J.~Schulman, and P.~Abbeel, ``Benchmarking deep reinforcement learning for continuous control,'' in \emph{ICML}, 2016, pp. 1329--1338.

\bibitem{williams2017information}
G.~Williams, N.~Wagener, B.~Goldfain, P.~Drews, J.~M. Rehg, B.~Boots, and E.~A. Theodorou, ``Information theoretic mpc for model-based reinforcement learning,'' in \emph{ICRA}.\hskip 1em plus 0.5em minus 0.4em\relax IEEE, 2017, pp. 1714--1721.

\bibitem{lee2021learning}
R.~Lee, D.~Ward, V.~Dasagi, A.~Cosgun, J.~Leitner, and P.~Corke, ``Learning arbitrary-goal fabric folding with one hour of real robot experience,'' in \emph{CoRL}, 2021, pp. 2317--2327.

\bibitem{mo2022foldsformer}
K.~Mo, C.~Xia, X.~Wang, Y.~Deng, X.~Gao, and B.~Liang, ``Foldsformer: Learning sequential multi-step cloth manipulation with space-time attention,'' \emph{RA-L}, vol.~8, no.~2, pp. 760--767, 2022.

\bibitem{longhini2023elastic}
A.~Longhini, M.~Moletta, A.~Reichlin, M.~C. Welle, A.~Kravberg, Y.~Wang, D.~Held, Z.~Erickson, and D.~Kragic, ``Elastic context: Encoding elasticity for data-driven models of textiles,'' in \emph{IEEE ICRA}, 2023, pp. 1764--1770.

\bibitem{longhini2022edo}
A.~Longhini, M.~Moletta, A.~Reichlin, M.~C. Welle, D.~Held, Z.~Erickson, and D.~Kragic, ``Edo-net: Learning elastic properties of deformable objects from graph dynamics,'' \emph{arXiv:2209.08996}, 2022.

\bibitem{liu2023grounding}
S.~Liu, Z.~Zeng, T.~Ren, F.~Li, H.~Zhang, J.~Yang, C.~Li, J.~Yang, H.~Su, J.~Zhu, \emph{et~al.}, ``Grounding dino: Marrying dino with grounded pre-training for open-set object detection,'' \emph{arXiv:2303.05499}, 2023.

\bibitem{kirillov2023segment}
A.~Kirillov, E.~Mintun, N.~Ravi, H.~Mao, C.~Rolland, L.~Gustafson, T.~Xiao, S.~Whitehead, A.~C. Berg, W.-Y. Lo, \emph{et~al.}, ``Segment anything,'' \emph{arXiv:2304.02643}, 2023.

\bibitem{scarselli2008graph}
F.~Scarselli, M.~Gori, A.~C. Tsoi, M.~Hagenbuchner, and G.~Monfardini, ``The graph neural network model,'' \emph{IEEE transactions on neural networks}, vol.~20, no.~1, pp. 61--80, 2008.

\bibitem{qi2017pointnet++}
C.~R. Qi, L.~Yi, H.~Su, and L.~J. Guibas, ``Pointnet++: Deep hierarchical feature learning on point sets in a metric space,'' \emph{NeurIPS}, vol.~30, 2017.

\bibitem{benelot2018}
B.~Ellenberger, ``Pybullet gymperium,'' \url{https://github.com/benelot/pybullet-gym}, 2018--2019.

\bibitem{garcia2024standard}
I.~Garcia-Camacho, A.~Longhini, M.~Welle, G.~Alenya, D.~Kragic, and J.~Borras, ``Standardization of cloth objects and its relevance in robotic manipulation,'' in \emph{IEEE ICRA}, 2024.

\bibitem{longhini2021textile}
A.~Longhini, M.~C. Welle, I.~Mitsioni, and D.~Kragic, ``Textile taxonomy and classification using pulling and twisting,'' in \emph{IEEE/RSJ IROS}, 2021, pp. 7564--7571.

\bibitem{lillicrap2015continuous}
T.~P. Lillicrap, J.~J. Hunt, A.~Pritzel, N.~Heess, T.~Erez, Y.~Tassa, D.~Silver, and D.~Wierstra, ``Continuous control with deep reinforcement learning,'' \emph{arXiv:1509.02971}, 2015.

\bibitem{blanco2023benchmarking}
D.~Blanco-Mulero, O.~Barbany, G.~Alcan, A.~Colom{\'e}, C.~Torras, and V.~Kyrki, ``Benchmarking the sim-to-real gap in cloth manipulation,'' \emph{arXiv:2310.09543}, 2023.

\end{thebibliography}
\bibliographystyle{IEEEtran}

\end{document}